%% file: iclr2025_conference.tex
\documentclass{article} 
\usepackage{iclr2025_conference,times}
\iclrfinalcopy
\input{math_commands.tex}

\usepackage{hyperref}
\usepackage{url}

\usepackage{multirow}
\usepackage[utf8]{inputenc} 
\usepackage[T1]{fontenc}    
\usepackage{hyperref}       
\usepackage{booktabs}       
\usepackage{amsfonts}       
\usepackage{nicefrac}       
\usepackage{microtype}      
\usepackage{xcolor}         
\newtheorem{definition}{Definition}
\newtheorem{theorem}{Theorem}
\usepackage{graphicx} 
\usepackage{amsmath}
\usepackage{wrapfig}
\usepackage{algorithm}
\usepackage{algorithmic}
\def\eg{\emph{e.g.}} 

\def\ie{\emph{i.e.}}

\def\etal{\emph{et al.}}

\AtEndPreamble{
    \usepackage[capitalize]{cleveref}
    \crefname{section}{Sec.}{Secs.}
    \Crefname{section}{Section}{Sections}
    \Crefname{table}{Table}{Tables}
    \crefname{table}{Tab.}{Tabs.}
}

\title{Fair Deepfake Detectors Can Generalize}

\author{%
Harry Cheng \\
National University of Singapore \\
\texttt{xaCheng1996@gmail.com} \\
\And
Ming-Hui Liu \\
Shandong University \\
\texttt{liuminghui@mail.sdu.edu.cn} \\
\And
Yangyang Guo\thanks{Corresponding author.} \\
National University of Singapore \\
\texttt{guoyang.eric@gmail.com } \\ \And
Tianyi Wang \\
National University of Singapore \\
\texttt{terry.ai.wang@gmail.com } \\ \And
Liqiang Nie \\
Harbin Institute of Technology (Shenzhen) \\
\texttt{nieliqiang@gmail.com } \\
\And
Mohan Kankanhalli \\
National University of Singapore \\
\texttt{mohan@comp.nus.edu.sg} \\
}
\begin{document}

\maketitle

\begin{abstract}
Deepfake detection models face two critical challenges: generalization to unseen manipulations and demographic fairness among population groups. However, existing approaches often demonstrate that these two objectives are inherently conflicting, revealing a trade-off between them.
In this paper, we, for the first time, uncover and formally define a causal relationship between fairness and generalization.
Building on the back-door adjustment, we show that controlling for confounders (data distribution and model capacity) enables improved generalization via fairness interventions.
Motivated by this insight, we propose Demographic Attribute-insensitive Intervention Detection (DAID), a plug-and-play framework composed of: i) Demographic-aware data rebalancing, which employs inverse-propensity weighting and subgroup-wise feature normalization to neutralize distributional biases; and ii) Demographic-agnostic feature aggregation, which uses a novel alignment loss to suppress sensitive-attribute signals.
Across three cross-domain benchmarks, DAID consistently achieves superior performance in both fairness and generalization compared to several state-of-the-art detectors, validating both its theoretical foundation and practical effectiveness.
\end{abstract}

\input{_Sec/1_intro}

\input{_Sec/2_related_work}
\input{_Sec/3_proof}
\input{_Sec/4_Method}
\input{_Sec/5_Exp}
\input{_Sec/6_conclusion}

\bibliography{iclr2025_conference}
\bibliographystyle{iclr2025_conference}


\end{document}

%% file: math_commands.tex
\usepackage{amsmath,amsfonts,bm}

\def\eqref#1{equation~\ref{#1}}

\def\1{\bm{1}}

\DeclareMathAlphabet{\mathsfit}{\encodingdefault}{\sfdefault}{m}{sl}
\SetMathAlphabet{\mathsfit}{bold}{\encodingdefault}{\sfdefault}{bx}{n}



%% file: _Sec/1_intro.tex
\section{Introduction}
\label{sec:intro}
With the advancement of cutting-edge facial synthesis models, attackers can generate high-quality forged faces at minimal cost~\cite{Region_aware_swapping, faceshifter}, resulting in serious negative social implications~\cite{wang2024deepfake}.
In response to these threats, numerous deepfake detection methods have been proposed~\cite{guan2024improving, li2018ictu, jointAV}. Employing binary real/fake classification~\cite{MAT, F3Net}, these approaches have achieved promising results when trained and tested on datasets with similar distributions (\ie, forged samples generated using the same manipulation techniques).
However, their generalization ability remains limited when faced with previously unseen forgery methods~\cite{x-ray, What_Makes_Fake, RFM, SRM, Self_ADV, Face_Reconstruction, cvpr_2025_yan, CVPR2025_1}.

\begin{figure}[t]
    \centering
    \includegraphics[width=0.9\textwidth]{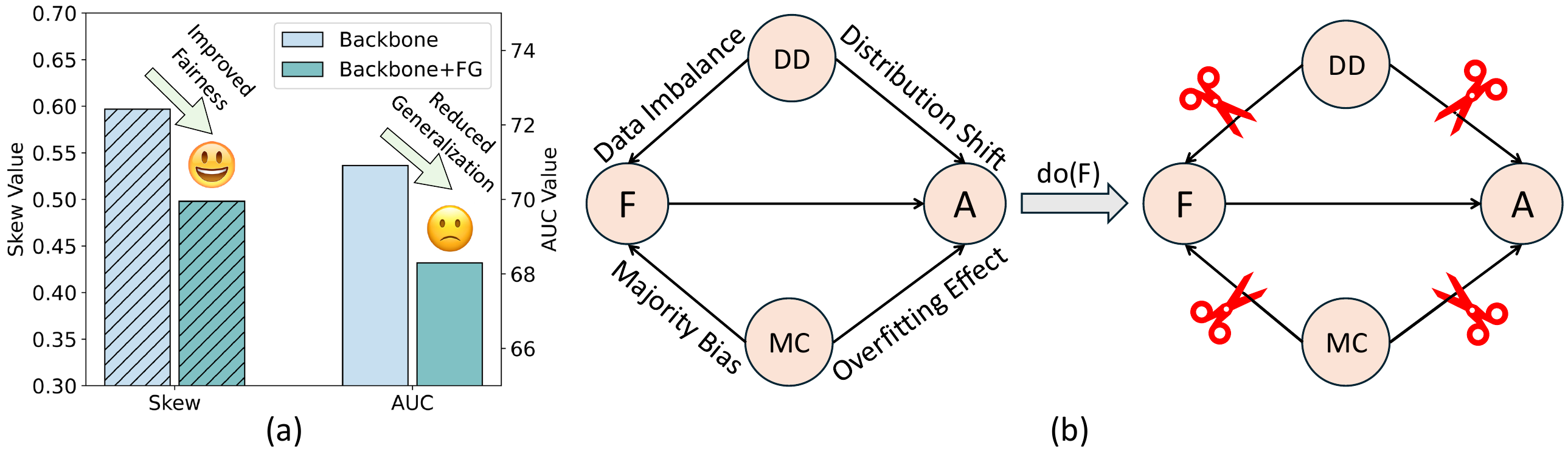}
    \vspace{-1em}
    \caption{(a) Comparison of model performance on Celeb-DF on Skew~\cite{Skew} (fairness metric, the lower the better) and AUC (generalization metric, the higher the better). FG~\cite{Fairness2024} is a method to improve fairness, but it may compromise the detector’s generalization ability. (b) Causal graph for relationship between fairness and generalization, where data distribution ($DD$) and model capacity ($MC$) act as confounders, \ie, they can affect both metrics, thereby obscuring the true causal relationship.}
    \vspace{-1em}
    \label{fig:intro}
\end{figure}

On the other hand, the fairness of deepfake detectors has also drawn increasing attention~\cite{FairAdapter_ICASSP, thinking_fairness_2025}. The problem lies in that a detector should maintain consistent performance across different demographic groups, such as gender and race. 
However, prior studies~\cite{gender_shae_fairness, correa2022systematic_fairness, lin2023improving_fairness} have predominantly shown that simply improving cross-domain generalization does not benefit all demographic subgroups equally (\ie, generalization $\!\not\rightarrow\!$ fairness).
Meanwhile, as shown in \Cref{fig:intro}a, pushing detectors to be more fair can compromise generalizability, which arguably makes these two a trade-off~\cite{fairness_wacv}. 


Different from existing studies that treat fairness and generalization as competing objectives, our preliminary experiments show that improving detector fairness can occasionally lead to enhanced cross-domain generalization. This finding motivates our hypothesis that demographic fairness causally improves generalization performance (\ie, fairness $\rightarrow$ generalization), although this effect is often obscured by confounders. 
To formalize this intuition, we construct a causal graph (see \Cref{fig:intro}b) in which fairness ($F$) functions as a treatment variable exerting a causal influence on generalization ($A$). However, data distribution ($DD$) and model capacity ($MC$) act as confounders affecting both metrics and potentially obscuring the true causal relationship.
To mitigate these confounding effects, we apply the back-door adjustment~\cite{Causality_book}, which blocks spurious paths and ensure that $A$ is influenced solely by $F$.
Specifically, we explicitly stratify the dataset based on human demographic attributes and control for model capacity (see \Cref{sec:proof} for details).
This procedure enables a rigorous estimation of the unbiased causal effect of fairness interventions on generalization performance across unseen manipulation methods.


To further validate our insight, we propose a novel Demographic Attribute-insensitive Intervention Detection (DAID) approach. 
Rather than directly optimizing for cross-domain generalization~\cite{SBI_ShioharaY22, UCF_0002ZFW23}, DAID explicitly control for both data distribution and model capacity confounders. 
In doing so, DAID elucidates the causal relationship between fairness and generalization during training, and generalization can be improved by intervening on fairness.
To this end, our DAID is equipped with two complementary modules. 
First, we apply a demographic-aware data rebalancing module, which uses adaptive sample reweighting and per-group normalization to mitigate distributional bias. 
Second, we propose demographic-agnostic feature aggregation, which aligns same-label samples across different demographic groups through a demographic-agnostic optimization strategy. 
Together, these modules serve distinct but synergistic purposes: the data rebalancing module ensures equitable representation across subgroups, while the feature aggregation module enhances the model’s ability to mitigate the influence of human-related attributes. As a result, DAID effectively controls both data- and model-level confounders, while achieving substantial improvements in fairness.


We conduct extensive experiments across multiple datasets and different backbones. The results demonstrate that our approach leads to improvements in both fairness and generalization. For instance, on the DFDC~\cite{dfdc}, DFD~\cite{dfd2019}, and Celeb-DF~\cite{Celeb-DF} datasets, our method outperforms several the state-of-the-art (SoTA) approaches. Our contributions are threefold:
\begin{itemize}
    \item 
    To the best of our knowledge, we are the first to establish a causal relationship where enhancing fairness leads to improved generalization in deepfake detection. This finding reveals a one-stone-hits-two-birds strategy: It enables the development of fairness-aware strategies that also enhance robustness.
    \item 
    We propose a novel approach that improves generalization by promoting fairness. Our method controls the confounders, thereby isolating the causal relationship between fairness and generalization and achieving improvement in both objectives.
    \item 
    We evaluate our approach on multiple datasets and backbones, showing consistent improvements in fairness and generalization. Code is provided in the supplementary materials.
\end{itemize}

%% file: _Sec/2_related_work.tex
\section{Related Work}
\subsection{Deepfake Detection}
\noindent\textbf{Generalization in Deepfake Detection.} Deepfake detection~\cite{Hong_Deepfake_CVPR_2024,CVPR24_Yan_Aug,xia2024mmnet,guan2024improving, liu2025data} is generally cast as a binary classification task. Preliminary efforts often endeavor to detect the specific manipulation traces~\cite{Exploring_Frequency_Adversarial, Two-Branch, SSTNET, pmlr-v235-zhang24aj}, which have shown certain improvements on intra-dataset setting. However, these methods often encounter inferior performance when applied to data with different distributions or manipulation methods. To address this generalization issue~\cite{Tan_CVPR24,li2023logical}, subsequent research has increasingly devoted efforts to learning more generalized features~\cite{CVPR2025_1, cvpr_2025_yan, liu2025learning}. For instance, RealForensics~\cite{Leveraging_Real_Talking} exploits the visual and auditory correspondence in real videos to enhance detection performance~\cite{VFD}. Shiohara~\etal~\cite{SBI_ShioharaY22} introduce a self-blended method to capture boundary-fusion features. 
Han~\etal~\cite{CVPR2025_1} apply facial component guidance to enhance spatial learning generalizability by encouraging the model to focus on key facial regions.

\noindent\textbf{Fairness in Deepfake Detection.}
Fairness in deepfake detection pertains to potential biases against certain demographic groups~\cite{Fairness_2021, fairness_2022, fairness_evaluation_2022}, particularly in terms of race and gender~\cite{fairness_2022_ICPR, FairAdapter_ICASSP}. For instance, Pu~\etal~\cite{fairness_evaluation_2022} evaluate the fairness of the detector MesoInception-4 and find it to be unfair to both genders. 
Some recent approaches~\cite{thinking_fairness_2025} have been proposed to address this problem by chasing for improved fairness metrics. 
For instance, Ju~\etal~\cite{fairness_wacv} mitigate sharp loss landscapes during training to improve fairness within the same data domain. Lin~\etal~\cite{Fairness2024} aims to enhance cross-domain fairness by leveraging contrastive learning across different demographic subgroups. 
Nevertheless, these methods treat fairness as the main optimization objective, without establishing a clear connection between fairness and generalization.

\subsection{Causality Inference}
In recent years, causal inference has emerged as a powerful tool to uncover causal relationships~\cite{overview_causal, causal_overview_2, causal_overview_3}. 
A growing body of research confirms that robust causal identification can lead to substantial improvements in model performance~\cite{Lv_2022_CVPR, mahajan2021domain, 10239469}.
Causal inference methods can be categorized into back-door and front-door adjustment~\cite{Pearl_causal, Pearl_causal_2}. The backdoor adjustment removes the confounding bias by stratifying the data according to the values of the confounders~\cite{zhang2020devlbert}. Li~\etal~\cite{23MM_causal} leverage back-door adjustment to mitigate inter- and intra-modal confounding, resulting in improved image-text matching accuracy. 
Chen~\etal~\cite{chen2023causal} apply back-door causal intervention to neutralize the textual bias to detect fake news. 
In contrast, the front door adjustment recovers the causal effect of a treatment by conditioning an observed mediator that fully carries the influence of the treatment on the outcome~\cite{chen2024mecd}.
For instance, Zhang~\etal~\cite{zhang2025causal} employ LLM-generated prompts as a mediator and calculate the causal effect between prompts and responses. In this paper, we apply back-door adjustment to block the influence of confounders, thus demonstrating the causal relationship between fairness and generalization.

%% file: _Sec/3_proof.tex
\section{Causal Analysis Between Fairness and Generalization}
\label{sec:proof}
\subsection{Causal Relationship Construction}
\noindent\textbf{Causal Graph.}
Figure~\ref{fig:intro}b illustrates our assumed causal structure as a directed acyclic graph (DAG) over four variables: fairness ($F$), generalization performance ($A$), data distribution ($DD$), and model capacity ($MC$). $F$ serves as a binary treatment variable: `low fairness' vs. `high fairness', based on the absolute value of Skew metric (smaller Skew indicates greater fairness). $A$ is the testing-set AUC, reflecting the generalization capability. $DD$ captures the distribution of sensitive attributes (\eg, race, gender), while $MC$ denotes the model’s architectural capacity. Since $DD$ and $MC$ influence both $F$ and $A$, we must control for them to isolate the causal effect of fairness on generalization.

This DAG contains two types of paths:
i) \textbf{Causal path}: $F\!\rightarrow\!A$ represents our hypothesis that improving fairness boosts generalization;
ii) \textbf{Confounding paths}: $DD \rightarrow \{F, A\}$, $MC \rightarrow \{F, A\}$, where data distribution and model capacity each affect both fairness and generalization.
Confounding paths that simultaneously influence both $F$ and $A$, such as $F \leftarrow DD \rightarrow A$ and $F\leftarrow MC\rightarrow A$, can induce a \emph{back-door effect}, introducing a spurious association between $F$ and $A$.

Therefore, it is essential to block these back-door effects for recovering the true causal effect of $F$ on $A$. To this end, we apply the \textbf{back-door adjustment}~\cite{Causality_book}. Specifically, if there exists a set of variables $\mathcal{Z}$ that satisfies the back-door criterion, we can estimate the causal relationship by conditioning on $\mathcal{Z}$. 
\begin{definition}[Back‑door Criterion]
Let $\mathcal{G}$ be a causal DAG and let $X$ and $Y$ be two nodes in $\mathcal{G}$. A set of variables $\mathcal{Z}$ satisfies the \emph{back‑door criterion} relative to $X, Y$ if:
\begin{enumerate}
  \item No element of $\mathcal{Z}$ is a descendant of $X \in G$.
  \item $\mathcal{Z}$ blocks every path between $X$ and $Y$ that begins with an arrow pointing into $X$.
\end{enumerate}
\vspace{-1em}
\end{definition}
In this study, $\mathcal{Z}$ is defined to include both the data and the model factors, \ie, $\mathcal{Z}$ = $\{DD, MC\}$. 

\begin{theorem}[Back‑door Adjustment Formula]
\label{theorem:the_1}
If a set $\mathcal{Z}$ satisfies the back‑door criterion relative to $X, Y$ in $\mathcal{G}$, then the causal effect of $X$ on $Y$ is identifiable and given by:
\begin{equation}
    \mathbb{P}\bigl(Y|{do}(X\!=\!x)\bigr)\!\!=\!\!\sum_{z} \mathbb{P} \bigl(Y|X\!=\!x, \mathcal{Z}\!=\!z\bigr)P(\mathcal{Z}\!=\!z).
\end{equation}
\vspace{-1em}
\end{theorem}
Here, ${do}(X\!=\!x)$ denotes an intervention that forcibly sets $X$ to $x$, disconnecting it from its natural causes. This allows us to distinguish causal effects from spurious associations in observational data.
Theorem~\ref{theorem:the_1} demonstrates that as long as the conditional distribution $\mathbb{P}(Y \mid X, \mathcal{Z})$ and the marginal distribution of the confounder set $ \mathbb{P}(\mathcal{Z})$ can be observed, the causal effect can be identified without experimental randomization. 
In our context, if the influence of varying fairness levels $F$ on generalization performance $A$ remains consistent when conditioned on different values of $DD$ and $MC$, then a direct causal relationship between fairness and generalization can be established.

\subsection{Causal Effect Estimation}

According to the back-door criterion, adjusting for
$\mathcal{Z}=\{DD,MC\}$\footnote{We approximate $\mathbb{P}(DD,MC)$ by the empirical frequency in the
\emph{held-out} test set, assuming that this set is an i.i.d.\ sample from the
deployment population.} suffices:
\begin{equation}
 \mathbb{P}\bigl(A\mid\operatorname{do}(F\!=\!f)\bigr)=
\sum_{dd,mc} \mathbb{P}\bigl(A\mid F\!=\!f,DD\!=\!dd,MC\!=\!mc\bigr)\,
           \mathbb{P}(DD\!=\!dd,MC\!=\!mc),
\label{eq:bd-formula}
\end{equation}
where $f$, $dd$, and $mc$ represent the values of $F$, $DD$, and $MC$, respectively. For simplicity, we discretize the two levels of fairness with a binary variable $\{0, 1\}$, where $f=0$ denotes low fairness. To examine the causal effect of $F$ on $A$, we define the Average Causal Effect (ACE) as follows:
\begin{equation}
\begin{aligned}
    \mathrm{ACE} &= \mathbb{P}\bigl(A\mid \mathrm{do}(F=1)\bigr)\;-\; \mathbb{P}\bigl(A\mid \mathrm{do}(F=0)\bigr) \\
    & = \sum_{dd,mc} \Bigl[\mathbb{P}(A\mid F=1,dd,mc) - \mathbb{P}(A\mid F=0,dd,mc)\Bigr]\;\mathbb{P}(dd,mc).\\
\end{aligned}
\end{equation}
In other words, the causal effect is defined as the weighted average of the performance differences observed between high and low fairness conditions within each subgroup. Moreover, we define $\mu_0 \;=\; \mathbb{P}\bigl(A\mid \mathrm{do}(F=0)\bigr)$, for any fairness level $f$, we can apply a simple substitution:
\begin{equation}
\begin{aligned}
\mathbb{P}\bigl(A\mid \mathrm{do}(F=f)\bigr)
&= \mu_0 \;+\; f\cdot\underbrace{\Bigl[\mathbb{P}\bigl(A\mid \mathrm{do}(F=1)\bigr) - \mathbb{P}\bigl(A\mid \mathrm{do}(F=0)\bigr)\Bigr]}_{\mathrm{ACE}}\\
&= \mu_0 + f\cdot\mathrm{ACE}.
\end{aligned}
\end{equation}
This leads to a straightforward linear formulation: When $f = 0$, we have $\mathbb{P}(A \mid \mathrm{do}(F = 0)) = \mu_0$. When $f = 1$, we have $\mathbb{P}(A \mid \mathrm{do}(F = 1)) = \mu_0 + \mathrm{ACE}$.
As long as $\mathrm{ACE} \neq 0$, we can assert that fairness $F$ has a causal effect on generalization performance $A$: $\mathrm{ACE} > 0$ implies that improving fairness leads to better model performance, and $\mathrm{ACE} < 0$ indicates the opposite.

We further design a concrete experiment to estimate the ACE to establish the causal relationship between fairness and generalization (more details are provided in the supplementary materials).

\noindent\textbf{Confounder Stratification.} 
For $DD$, we stratify the dataset based on the intersection of gender and race. Specifically, the dataset is first divided into two groups according to binary gender: Male and Female. Within each gender group, samples are further categorized by skin tone into three subgroups: White, Black, and Asian. Each intersection of gender and race is treated as a distinct demographic distribution.
For $MC$, we employ two different architectures: Xception~\cite{Xception} (lower capacity) and EfficientNet~\cite{Efficient} (higher capacity), the latter of which is known for stronger cross-domain performance~\cite{CVPR24_Yan_Aug}.

\noindent\textbf{Fairness Intervention ($do(F)$).}
We implement two training regimes to approximate $do(F=0)$ and $do(F=1)$~\cite{Do-cal}: 1) Low fairness ($F=0$): Standard cross-entropy training. 2) High fairness ($F=1$): Cross-entropy loss with a simple resampling strategy~\cite{cheng2024social}, where each sample in the cross-entropy loss is assigned a weight to suppress the over-representation of majority groups.

\noindent\textbf{ACE Estimation Results.} Based on the above procedure, we observe an average ACE gain of \textbf{2.35 percentage points}
(stratified bootstrap resampling with B = 1000, $\Delta$ = 0.0235, 95\% CI [0.0186, 0.0280], two-sided $p < 0.001$).
This result indicates that, after removing the influence of confounders, a direct relationship between fairness and generalization emerges.

%% file: _Sec/4_Method.tex
\subsection{Demographic Attribute-Insensitive Intervention Detection}
\label{sec:method}
Motivated by our causal findings, we conclude that, as long as confounders are properly controlled, the clear causal pathway can be leveraged to enhance generalization by intervening on more readily measurable fairness.
Therefore, we introduce Demographic Attribute-Insensitive Intervention Detection (DAID), a training approach that uses fairness interventions to boost cross-domain generalization. 

\begin{figure}
    \centering
    \includegraphics[width=0.90\linewidth]{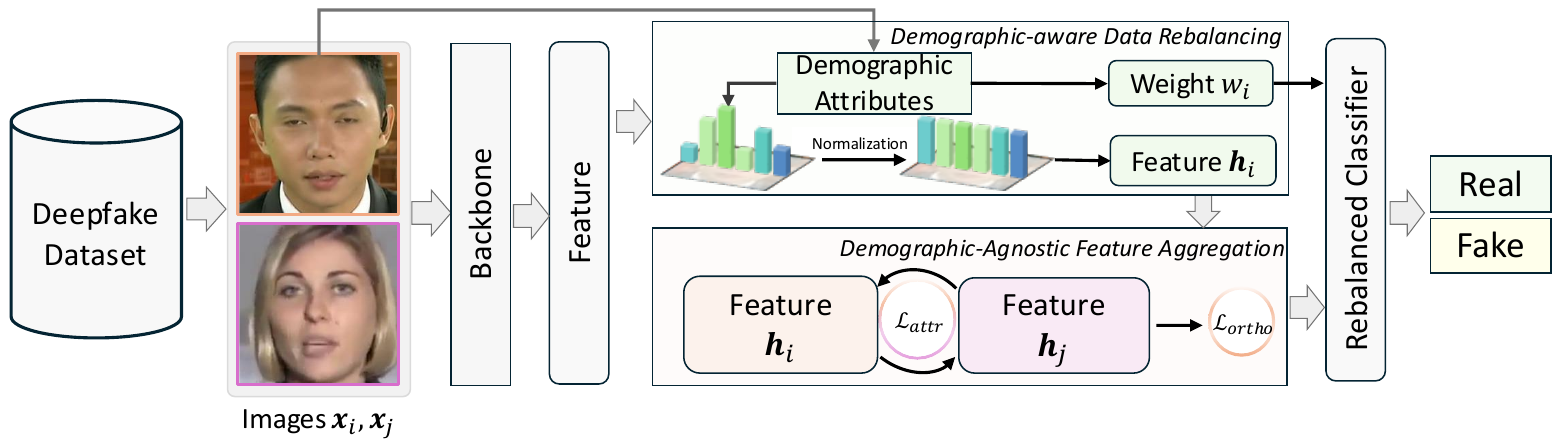}
    \vspace{-0.5em}
    \caption{Overview of the proposed DAID method. \textbf{Top}: Demographic-aware Data Rebalancing. We utilize human attributes to perform demographic normalization and classifier rebalancing, which suppresses the confounding effects of $DD$. \textbf{Bottom}: Demographic-Agnostic Feature Aggregation. We introduce a demographic-agnostic loss that enhances the model’s ability to filter out demographic-related information, which mitigates the confounding influence of $MC$ while improving fairness.}
    \label{fig:pipeline}
    \vspace{-1em}
\end{figure}

As illustrated in \Cref{fig:pipeline}, DAID counteracts two key confounders: data distribution ($DD$) and model capacity ($MC$) via two complementary modules: 
i) Demographic-aware Data Rebalancing, and ii) Demographic-Agnostic Feature Aggregation.




\noindent\textbf{Demographic-aware Data Rebalancing.}
To neutralize the spurious dependency induced by the data distribution confounder $DD$, our rebalancing module includes two key components: sample-wise reweighting and representation-level normalization, that jointly calibrate both the optimization direction and the feature space geometry~\cite{fairness_norm}.

Firstly, we employ the inverse‐probability reweighting strategy. 
Let $\mathbf{x}_i$ denote an input sample with sensitive demographic attributes $\mathbf{s}_i$ (\eg, gender, race). To equalize the influence of majority and minority groups, we compute a sample-specific importance weight:
\begin{equation}
w_i = \left( \prod_{k=1}^{K} \widehat{\mathbb{P}}\big(\mathbf{s}_i^{(k)}\big) \right)^{-1},
\label{eqn:weight}
\end{equation}
where $s_i^{(k)}$ is the $k$-th sensitive attribute of $\mathbf{x}_i$, and $\widehat{\mathbb{P}}\big(s_i^{(k)}\big)$ is the empirical marginal frequency estimated from the training data. This inverse propensity weighting ensures that the expected contribution of each demographic subgroup to the loss function is approximately uniform, thus suppressing spurious correlations between $DD$ and the optimization target.

Beyond reweighting, we further mitigate $DD$-induced feature shifts by normalizing latent features within each subgroup. Denote the feature vector for $\mathbf{x}_i$ as $\mathbf{h}_i$. For each $DD$ group $dd$, we estimate the first and second moments:
\begin{equation}
\boldsymbol{\mu}_{dd} = \mathbb{E}_{i: dd_i = dd}[\mathbf{h}_i], \quad
\boldsymbol{\sigma}^2_{dd} = \mathrm{Var}_{i: dd_i = dd}[\mathbf{h}_i],
\end{equation}
and apply the following demographic-conditioned normalization:
\begin{equation}
\hat{\mathbf{h}}_i = \frac{\mathbf{h}_i - \boldsymbol{\mu}_{dd_i}}{\sqrt{\boldsymbol{\sigma}^2_{dd_i} + \varepsilon}}.
\end{equation}
This operation aligns the group-conditioned feature distributions, removing systematic shifts induced by demographic imbalance and restoring feature comparability across subgroups.

In summary, these two strategies decouple the confounding influence of $DD$ from both model updates and representation space, yielding unbiased learning that better reflect the intrinsic relationship between fairness ($F$) and generalization ($A$).

\noindent\textbf{Demographic-Agnostic Feature Aggregation.}
To eliminate the confounding influence of $MC$, we propose to encourage the model to focus on task-relevant cues while marginalizing residual demographic signals. Therefore, we perform demographic-invariant optimization in the learned representation space. The key intuition is that manipulation-consistent samples, \ie, those with the same class label but differing sensitive attributes, should lead to similar internal representations. 

Formally, let $\mathcal{P} = \{(\mathbf{x}_i, \mathbf{x}_j)\}$ be a set of sample pairs such that $y_i = y_j$ (same task label) and $dd_i \neq dd_j$ (different demographic attributes). We enforce:
\begin{equation}
\mathcal{L}_{\mathrm{attr}} = \frac{1}{|\mathcal{P}|} \sum_{(i,j) \in \mathcal{P}} \mathcal{L}_{\mathrm{cos}}(\hat{\mathbf{h}}_i, \hat{\mathbf{h}}_j),
\label{eqn:attr}
\end{equation}
where $\hat{\mathbf{h}}_i$ and $\hat{\mathbf{h}}_j$ are normalized feature vectors, and $\mathcal{L}_{\mathrm{cos}}(\cdot, \cdot)$ denotes a cosine similarity loss:
\begin{equation}
    \mathcal{L}_{\cos}\left(\mathbf{h}_i, \mathbf{h}_j\right)=1-\cos \left(\mathbf{h}_i, \mathbf{h}_j\right)+\epsilon
\end{equation}
where $\mathrm{cos}(\cdot)$ denotes the cosine similarity between feature vectors. To ensure this alignment occurs in a semantically meaningful subspace, we factorize $\hat{\mathbf{h}} \in \mathbb{R}^d$ via a low-rank projection layer:
\begin{equation}
\tilde{\mathbf{h}} = \mathbf{U}^\top \hat{\mathbf{h}},
\end{equation}
where $\mathbf{U}$ is a trainable orthonormal basis, used to filter out irrelevant directions. To avoid collapsing to trivial solutions, we regularize the projected features with:
\begin{equation}
\mathcal{L}_{\mathrm{ortho}} = \| \mathbf{U} \mathbf{U}^{\top} - \mathbf{I} \|^2_F,
\label{eqn:svd}
\end{equation}
where $\mathbf{I}$ is the identity matrix, and $|\cdot|_F$ denotes the Frobenius norm.

By enforcing demographic-invariant structure in a filtered representation space, this module suppresses the model’s reliance on demographic features, thereby neutralizing $MC$ as a confounder and sharpening the causal interpretability of fairness-driven generalization.

\noindent\textbf{Training Objective.}
We adopt a fully end-to-end optimization strategy that preserves the backbone architecture of the base detector. Specifically, we only insert our proposed modules before the classification head. It worth noting that our approach is model-agnostic and can be seamlessly integrated into various deepfake detection backbones, which ensures inference efficiency.

Let $f_\theta: \mathbf{x} \mapsto \mathbf{h}$ denote the backbone encoder, and $g_\phi: \mathbf{h} \mapsto \hat{y}$ denote the binary classifier. Our total objective integrates the classification loss with two fairness-enhancing regularizers:
\begin{equation}
\mathcal{L}_{\mathrm{total}} = \mathcal{L}_{\mathrm{cls}} + \lambda_{\mathrm{attr}} \mathcal{L}_{\mathrm{attr}} + \lambda_{\mathrm{ortho}}\mathcal{L}_{\mathrm{ortho}},
\end{equation}
where $\mathcal{L}_{\mathrm{cls}} = \mathbb{E}{(\mathbf{x}, y)}\big[w_i \cdot \mathcal{B}\big(g_\phi(f_\theta(\mathbf{x})), y\big)\big]$ is the weighted binary cross-entropy loss over labels and sample-specific importance weight (see \Cref{eqn:weight}); $\mathcal{L}_{\mathrm{attr}}$ enforces demographic-invariant alignment between same-label samples across subgroups (see \Cref{eqn:attr}); $\mathcal{L}_{\mathrm{ortho}}$ ensures that the projected representation remains compact and expressive (see \Cref{eqn:svd}). $\lambda_{\mathrm{attr}}, \lambda_{\mathrm{ortho}}$ are hyperparameters that modulate the contribution of each loss.


%% file: _Sec/5_Exp.tex
\section{Experiments}
\subsection{Datasets and Metrics}
\noindent\textbf{Datasets.}
Following prior work~\cite{CVPR24_Yan_Aug, sun2024rethinking, sun2025towards}, we employed FaceForensics++ (FF++) as the training set and evaluate the generalization performance on three other datasets: DFDC~\cite{dfdc}, DFD~\cite{dfd2019}, and Celeb-DF~\cite{Celeb-DF}. Since none of these datasets contain native demographic annotations, we follow the data processing, annotation protocol, and sensitive attribute intersection strategy of previous fairness studies~\cite{Fairness2024, FFpp_anno, fairness_wacv}. Specifically, we annotated each face with a combination of gender and race attributes, resulting in six demographic subgroups: Male-Asian (M-A), Male-White (M-W), Male-Black (M-B), Female-Asian (F-A), Female-White (F-W), and Female-Black (F-B).

\noindent\textbf{Metrics.}
We used AUC as the primary metric to evaluate the generalizability of the model and adopted Skew as the fairness metric~\cite{Skew, bias_vision_cvpr_20, cheng2024social}. Skew is a commonly used indicator for measuring model fairness, which quantifies the performance disparity across different demographic subgroups. In our context, a lower Skew value indicates better fairness, with Skew = 0 representing perfectly fair predictions.
The detailed computation of Skew is provided in the supplementary materials.

\subsection{Implementation details} 
We used several deepfake detectors as backbone models, including Xception~\cite{Xception}, F$^3$-Net~\cite{F3Net}, EfficientNet~\cite{Efficient}, and CADDM~\cite{CADDM}, to evaluate the effectiveness of DAID. Training employs AdamW (lr $1\times10^{-3}$, weight decay $4\times10^{-3}$) until convergence, with a batch size of 64. 
Images are resized to $224\times224$ and normalized by ImageNet statistics. All runs use a single H100 GPU.

\subsection{Main Results}
\begin{table}[t]
\centering
\scalebox{0.78}{
\begin{tabular}{l|l|cc|cc|cc}
\toprule \midrule
\multirow{2}{*}{Method} & \multicolumn{1}{c}{\multirow{2}{*}{Venue}} & \multicolumn{2}{c|}{DFDC} & \multicolumn{2}{c|}{DFD} & \multicolumn{2}{c}{Celeb-DF} \\ \cmidrule{3-8} 
                        &                        & Skew~$\downarrow$          & AUC~$\uparrow$      & Skew~$\downarrow$       & AUC~$\uparrow$        & Skew~$\downarrow$          & AUC~$\uparrow$          \\ \midrule
Xception~\cite{Xception}                 & ICCV'19                 & 2.221  & 60.63    & 0.564      & 80.69      &  0.597        & 70.91        \\
EffcientNet~\cite{Efficient}             & ICML'19                 & 2.011  & 60.49    & \underline{0.351}      & 83.12      &  0.437        & 75.36        \\ 
F$^3$-Net~\cite{F3Net}                   & ECCV'20                 & 2.143  & 60.17    & 0.589      & 77.68      &  0.556        & 74.36       \\ 
Face X-ray~\cite{x-ray}                  & CVPR'20                 & 1.982  & 62.00    & 0.821      & 80.46      &  0.491        & 74.20        \\ 
SBI~\cite{SBI_ShioharaY22}               & CVPR'22                 & 2.385  & 63.39    & 0.757      & 86.43      &  0.715        & 79.76        \\ 
RECCE~\cite{Face_Reconstruction}         & CVPR'22                 & 2.622  & 61.63    & 0.738      & 80.13      &  0.644        & 70.55        \\ 
GRU~\cite{GRU}                           & CVPR’24                 & 2.432  & 62.63    & 0.551      & 86.48      &  0.405        & 76.00        \\ 
CADDM~\cite{CADDM}                       & CVPR'23                 & 2.183  & 63.77    & 0.547      & 88.59      &  \underline{0.391}        & 81.75        \\ 
UCF~\cite{UCF_0002ZFW23}                 & CVPR'23                 & 2.272  & 60.03    & 0.510      & 81.01      &  0.619        & 71.73        \\ 
ProDet~\cite{Prodet}                     & NeurIPS'24              & 2.306  & \underline{65.89}    & 0.432      & 89.18      &  0.569        & \underline{82.71}         \\ 
VLFFD~\cite{sun2025towards}              & CVPR'25                 & 2.411  & 65.21    & 0.669      & \underline{90.08}      &  0.526        & 81.17        \\ \midrule
$^\ddag$DAW-FDD~\cite{fairness_wacv}      & WACV'24          & 2.127  & 59.96    & 0.528      & 71.40      &  0.509        & 69.55        \\
$^\ddag$FG~\cite{Fairness2024}            & CVPR'24          & \underline{1.932}  & 60.11    & 0.447      & 80.42      &  0.498        & 68.30   \\ \midrule
 DAID        & \multicolumn{1}{c|}{-}      & \textbf{1.460}  & \textbf{66.85}    & \textbf{0.263}    & \textbf{91.15}      &  \textbf{0.289}        & \textbf{84.39}        \\ \midrule \bottomrule
\end{tabular}}
\vspace{0.5em}
\caption{Frame-level cross-dataset performance comparison on fairness and generalization of baselines and our approach. We reproduced all baselines on three datasets and reported their Skew and AUC values.
$^\ddag$: This method is proposed to enhance the fairness of the detector.}
\vspace{-1.5em}
\label{tab:main_result}
\end{table}
In \Cref{tab:main_result}, we reported a comparison of our method, DAID, against several SoTA baselines in terms of both fairness and generalization performance. It can be seen that DAID consistently achieves the best results in all three datasets. For instance, on Celeb-DF, our method improves fairness by 26\% compared to the best-performing baseline. On the DFDC and DFD datasets, DAID achieves AUC scores of 66.85\% and 91.15\%, outperforming all competing methods. By controlling for confounding factors, we successfully achieve simultaneous improvements in both fairness and generalization.

It can be observed that achieving a high AUC does not necessarily imply high fairness. For example, VLFFD attains an AUC of 90.08\% on the DFD dataset.
However, its fairness performance lagged behind that of UCF, which exhibits significantly lower generalizability than VLFFD but demonstrates better fairness as indicated by a lower skew.
Moreover, fairness-oriented methods, \ie, DAW-FDD and FG, effectively enhance the fairness of the model. Nevertheless, this improvement may come at the cost of reduced generalization. For instance, on the Celeb-DF dataset, FG outperforms most baselines in terms of fairness, yet its AUC score is only around 68\%, significantly lower than those achieved by other methods.


\subsection{Ablation Studies}

\subsubsection{Comparison on Modules}
\begin{table}[]
\centering
\scalebox{0.78}{
\begin{tabular}{cc|cc|cccccc}
\toprule \midrule
\multicolumn{4}{c|}{Module}                                                     & \multicolumn{6}{c}{Dataset}                                                       \\ \midrule
\multicolumn{2}{c|}{Data Rebalancing} & \multicolumn{2}{c|}{Feature Aggregation}    & \multicolumn{2}{c}{DFDC} & \multicolumn{2}{c}{DFD} & \multicolumn{2}{c}{Celeb-DF} \\ \cmidrule{1-4} \cmidrule{5-10} 
Reweight           & Normalization   & \multicolumn{1}{c}{$\mathcal{L}_\mathrm{{attr}}$} & \multicolumn{1}{c|}{$\mathcal{L}_\mathrm{{ortho}}$}  & Skew~$\downarrow$          & AUC~$\uparrow$      & Skew~$\downarrow$       & AUC~$\uparrow$        & Skew~$\downarrow$          & AUC~$\uparrow$          \\ \midrule
-             &  -             &    -           &   -                           & 2.183         & 63.77         & 0.547         & 88.59         &  0.391        & 81.75       \\ \midrule
$\checkmark$  &                &                &                               & 1.719         & 64.94         & 0.295         & 89.63         &  0.340        & 83.07        \\ 
$\checkmark$  & $\checkmark$   &                &                               & 1.574         & 65.96         & 0.274         & 90.67         &  0.319        & 83.98        \\ 
              &                &$\checkmark$    &                               & 1.750         & 65.40         & 0.273         & 89.38         &  0.327        & 83.59        \\ 
              &                &$\checkmark$    &$\checkmark$                   & 1.715         & 64.96         & 0.271         & 89.55         &  0.321        & 83.88  \\  
$\checkmark$  & $\checkmark$   &$\checkmark$    &                               & 1.495         & 66.49         & 0.266         & 91.05         &  0.292        & 84.12  \\  
$\checkmark$  & $\checkmark$  &$\checkmark$     &$\checkmark$        &\textbf{1.460} &\textbf{66.85} &\textbf{0.263} &\textbf{91.15} &\textbf{0.289} & \textbf{84.39}  \\ \midrule 
\bottomrule
\end{tabular}
}
\vspace{0.5em}
\caption{Performance of ablation studies on each module of DAID.}
\vspace{-1em}
\label{tab:ablation}
\end{table}
We reported the ablation studies on the modules of our DAID in \Cref{tab:ablation}. Specifically, we incrementally integrate each DAID module into the backbone model to assess their individual contributions. The results indicate that omitting any single module negatively impacts performance. For instance, removing the data rebalancing module, \ie, no longer controlling the confounding factor $DD$, leads to a significant performance drop across all three datasets. Overall, the integration of all DAID modules yields the best performance in both generalization and fairness.

\subsubsection{Comparison on Hyperparameters}
\setlength{\columnsep}{5pt}   
\setlength{\intextsep}{5pt}   
\begin{wrapfigure}{r}{0.55\textwidth}  
  \vspace{-1em}
  \begin{center}
    \includegraphics[width=0.55\textwidth]{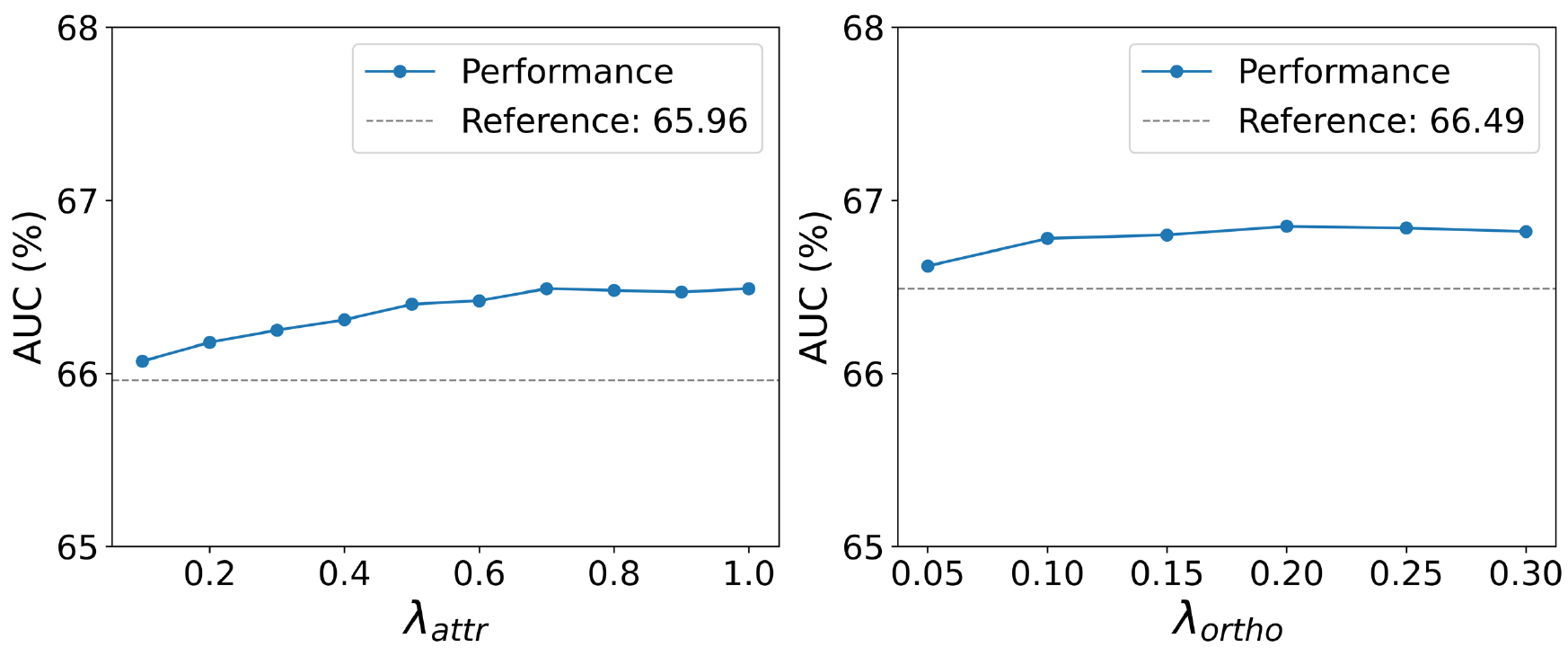}  
  \end{center}
  \vspace{-10pt}
  \caption{Hyperparameter analysis.}
  \label{fig:parameter}
  \vspace{-3pt}
\end{wrapfigure}
We employ two hyperparameters, $\lambda_{\mathrm{attr}}$ and $\lambda_{\mathrm{ortho}}$, to control the relative weights of the corresponding loss functions. To investigate their impact on model generalization, we conducted a parameter sensitivity analysis, with the results shown in \Cref{fig:parameter}. As both parameters increase, model performance initially improves and then stabilizes. Based on empirical observations, we select $\lambda_{\mathrm{attr}}$ = 0.7 and $\lambda_{\mathrm{ortho}}$ = 0.2 as default values. It worth noting that our method demonstrates robustness to hyperparameter selection. 

\subsubsection{Comparison on Demographic Attributes}
\begin{wrapfigure}{r}{0.50\textwidth}  
  \vspace{-1em}
  \begin{center}
    \includegraphics[width=0.49\textwidth]{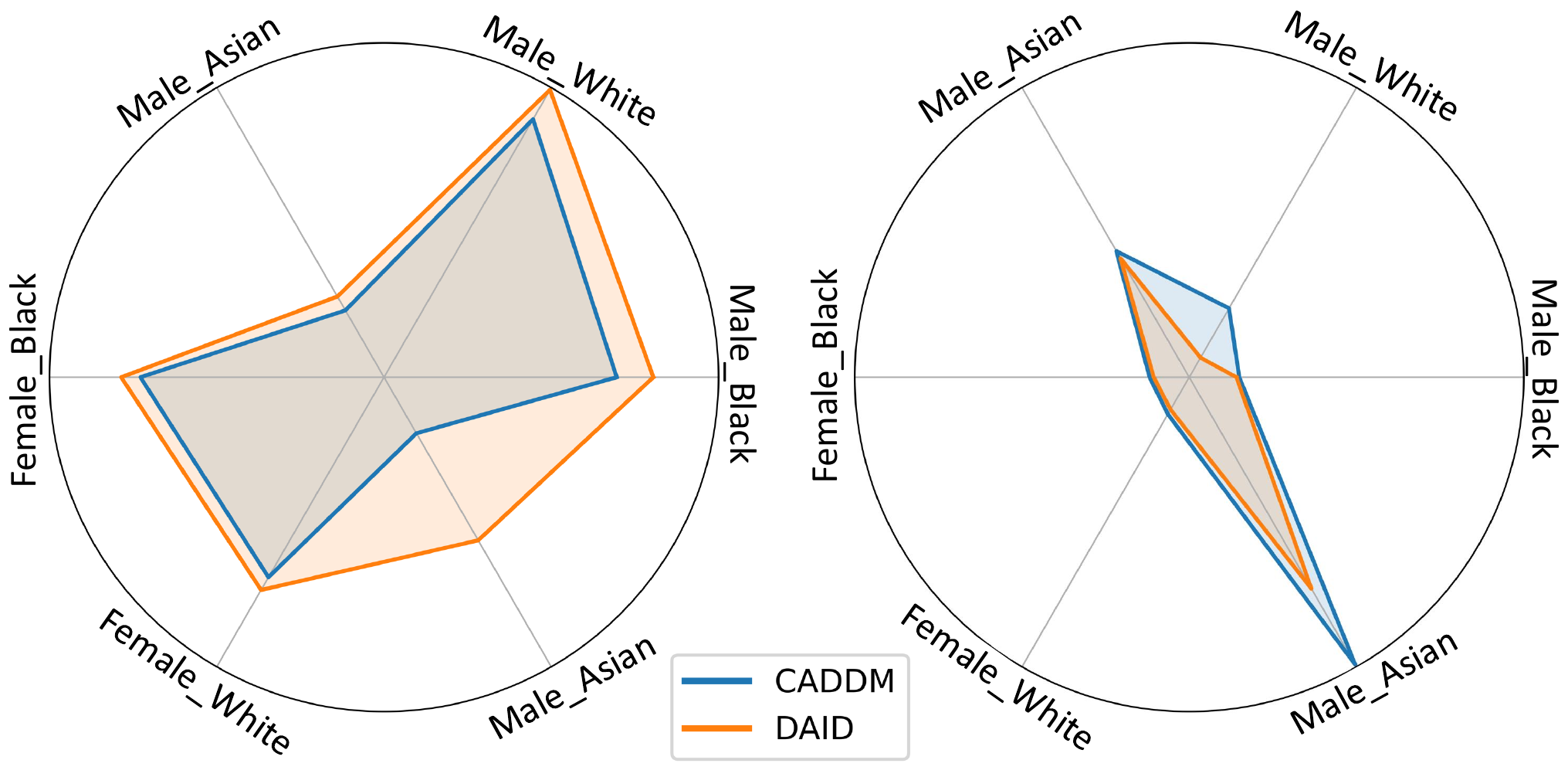}  
  \end{center}
  \vspace{-10pt}
  \caption{Radar plot for DAID. Left: AUC$\uparrow$ (\%) for generalization. Right: Skew$\downarrow$ for fairness.}
  \label{fig:radar}
  \vspace{-3pt}
\end{wrapfigure}
In \Cref{fig:radar}, we reported a radar plot that illustrates the performance of the model on the DFDC dataset at different intersections between gender and race, \eg, White-Female. The left subfigure presents the AUC performance for evaluating generalization. Our DAID model outperforms the baseline across all six demographic intersections, with particularly notable improvement on the Male-Asian subgroup, where AUC increases by 30\%. The right subfigure assesses fairness via the Skew metric, where our model demonstrates significantly lower skew values. This indicates that DAID achieves greater fairness in various demographic dimensions.

\subsubsection{Comparison on Backbones}
\Cref{tab:backbones} presents the performance when applying the DAID to different backbone architectures. Specifically, we compare the performance of the four backbones, \ie, Xception, EfficientNet, F$^3$-Net, and CADDM. 
As shown in the table, our method consistently enhances both fairness and generalization across all backbones. For instance, on Celeb-DF, applying our DAID to the Xception backbone yields a 5\% increase in AUC and nearly a 20\% improvement in fairness. It worth noting that this process does not require any architectural modifications to the model, leading to synergistic gains greater than the sum of individual improvements.

\begin{table}[]
   \centering
    \scalebox{0.75}{
    \begin{tabular}{l|cc|cccccc}
    \toprule \midrule
    \multirow{2}{*}{Method}  & \multicolumn{2}{c|}{FF++} & \multicolumn{2}{c}{DFDC} & \multicolumn{2}{c}{DFD} & \multicolumn{2}{c}{Celeb-DF} \\ \cmidrule{2-9} 
                              & Skew~$\downarrow$          & AUC~$\uparrow$      & Skew~$\downarrow$       & AUC~$\uparrow$       & Skew~$\downarrow$          & AUC~$\uparrow$      & Skew~$\downarrow$       & AUC~$\uparrow$       \\ \midrule
    Xception~\cite{Xception}                 & 0.177  & 97.85    & 2.221  & 60.63    & 0.564      & 80.69      &  0.597        & 70.91              \\
    +DAID                                    & \textbf{0.122}  & \textbf{98.64}    & \textbf{1.772}  & \textbf{63.36}  & \textbf{0.398}  & \textbf{82.54} & \textbf{0.467} & \textbf{75.23}        \\ \midrule
    EffcientNet~\cite{Efficient}             & 0.185  & 98.08    & 2.011  & 60.49   & 0.351      & 83.12      &  0.437        & 75.36           \\ 
    +DAID                                    & \textbf{0.136}  & \textbf{98.72}    & \textbf{1.697}  & \textbf{63.43}  & \textbf{0.264} & \textbf{84.31} & \textbf{0.352} & \textbf{78.49}        \\ \midrule
    F$^3$-Net~\cite{F3Net}                   & 0.219  & 97.32     & 2.143  & 60.17    & 0.589      & 77.68      &  0.556        & 74.36              \\ 
    +DAID                                    & \textbf{0.127}  & \textbf{97.63}    & \textbf{1.544}  & \textbf{62.68}  & \textbf{0.220} & \textbf{78.53} & \textbf{0.541} & \textbf{76.54}        \\ \midrule
    CADDM~\cite{CADDM}                       & 0.220  & 99.15    & 2.183  & 63.77    & 0.547      & 88.59      &  0.391        & 81.75            \\ 
    +DAID                                    & \textbf{0.119}  & \textbf{99.26}    & \textbf{1.460}  & \textbf{66.85}   & \textbf{0.263}    & \textbf{91.15}      &  \textbf{0.289}        & \textbf{84.39}       \\ \midrule \bottomrule
    \end{tabular}}
    \vspace{0.5em}
    \caption{Performance comparison after applying our DAID to different backbones. All models are trained on the FF++ dataset and evaluated on four datasets. Our method consistently leads to significant improvements across all backbone architectures.}
    \label{tab:backbones}
    \vspace{-1.5em}
\end{table}

\subsection{Visualization Results}
\begin{figure}
    \centering
    \includegraphics[width=0.95\linewidth]{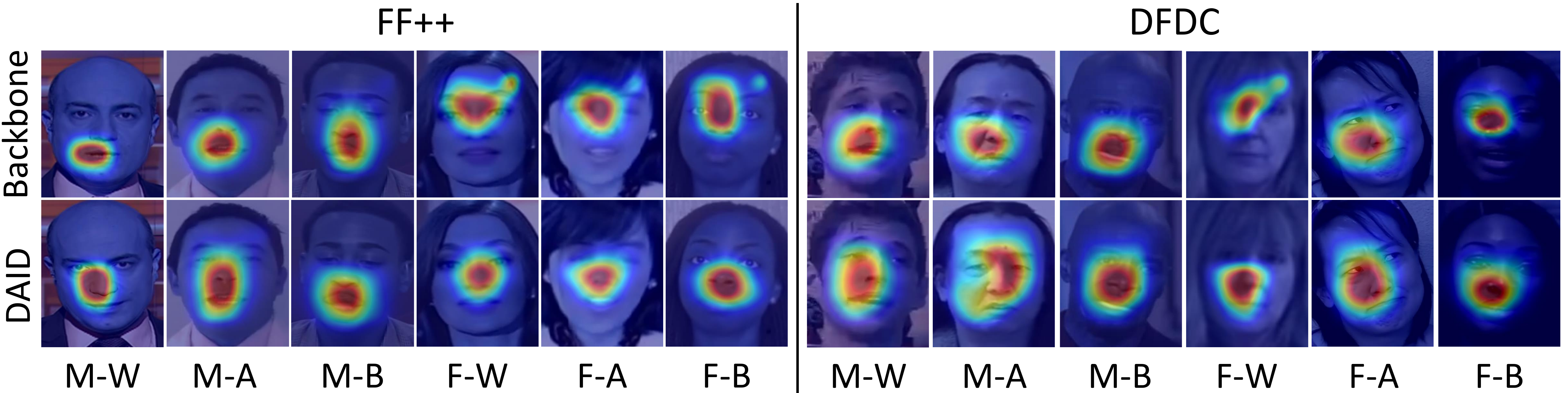}
    \vspace{-0.5em}
    \caption{Non-cherry-picked Heatmaps. We included heatmaps for six demographic subgroups across two datasets: Male-Asian (M-A), Male-White (M-W), Male-Black (M-B), Female-Asian (F-A), Female-White (F-W), and Female-Black (F-B).}
    \vspace{-1.5em}
    \label{fig:vis}
\end{figure}
In \Cref{fig:vis}, we present the heatmap results of the backbone model without fairness enhancement and our proposed DAID method. It can be seen that the backbone exhibits markedly different attention regions for different attributes. For instance, it focuses primarily on the lips for male subjects, while emphasizing the upper faces for female subjects. Furthermore, within the same gender, subtle differences in attention regions are also observed across different racial groups. For example, the backbone tends to focus more on the left side of the lips for the Male-White group, whereas for the Male-Black group, the nose is more frequently included in the attention region. This indicates that the backbone model conflates demographic attributes with cues for deepfake detection, potentially undermining reliable decision-making. In contrast, DAID demonstrates consistent detection patterns across both gender and race groups, effectively indicating that our method is insensitive to demographic attributes. Moreover, compared to the backbone, DAID generally focuses on broader regions of the image, reflected in its superior generalization capability.

\subsection{Efficiency Analysis}
We assess the additional computation introduced by DAID’s two modules on a single NVIDIA H100 GPU (batch size 64, input resolution 224×224).
For the data rebalancing module, the reweighting step adjusts only the classification loss based on subgroup frequencies, and subgroup-wise feature normalization operates directly on batch statistics. Neither requires extra gradient computations beyond standard training, resulting in negligible run-time impact.
For feature aggregation module, we introduce two regularization losses and a low-rank projection layer. These involve only light matrix multiplications and loss evaluations, resulting in minimal extra cost.
On EfficientNet, standard training takes 233 min for the full session. Incorporating DAID increases this to 243 min - a relative overhead of 4.3\%. Therefore, DAID’s fairness-driven interventions add under 5\% to total training time, making the framework practical for large-scale use.

%% file: _Sec/6_conclusion.tex
\section{Conclusion and Discussion}
In this paper, we demonstrate that improving fairness can causally lead to a better generalization in deepfake detection.
Building on this insight, We propose the Demographic Attribute-insensitive Intervention Detection (DAID), a novel plug-and-play approach that jointly ensures demographic fairness and generalization without modifying base architectures. Extensive experiments on various benchmarks validate the theoretical foundation and practical value of DAID. Our findings reframe fairness from a mere ethical concern into a strategic lever for enhancing model robustness. By harnessing fairness as a means to improve generalization, we offer a new perspective and a practical path toward building more robust and equitable deepfake detectors. However, one limitation of our current framework is its reliance on demographic annotations. Extending DAID to operate under unlabeled or multi-dimensional fairness settings remains an important direction for future work.